\documentclass[10pt,twocolumn,letterpaper]{article}
\usepackage[pagenumbers]{cvpr}
\def\confName{CVPR}
\def\confYear{2027}
\title{MoVelo: World-Grounded Human Motion Recovery with Feedback from Conditional Velocity Field}
\newcommand{\moveloinstitution}[3]{$^{#1}$\href{#2}{\textcolor{black}{#3}}}
\author{%
  Da Li$^{1}$ \quad Xunshu Yang$^{2}$ \quad Jiping Jin$^{3}$ \quad Ming Li$^{4}$ \quad Ce Zheng$^{2}$\thanks{Corresponding author.} \\[.75em]
  \moveloinstitution{1}{https://www.szu.edu.cn/}{Shenzhen University} \quad
  \moveloinstitution{2}{https://www.hust.edu.cn/}{Huazhong University of Science and Technology} \\
  \moveloinstitution{3}{https://www.auckland.ac.nz/}{University of Auckland} \quad
  \moveloinstitution{4}{https://www.cuhk.edu.cn/}{The Chinese University of Hong Kong, Shenzhen}%
}
\usepackage{bm}
\usepackage{float}
\usepackage{placeins}
\usepackage{needspace}
\usepackage{tabularx}
\usepackage{multirow}
\usepackage{colortbl}
\usepackage{pifont}
\usepackage{xurl}
\usepackage{microtype}
\definecolor{tablehighlight}{gray}{0.94}
\newcommand{\papertablestyle}{%
  \small
  \setlength{\tabcolsep}{3pt}%
  \renewcommand{\arraystretch}{1.0}%
}
\definecolor{cvprblue}{rgb}{0.21,0.49,0.74}
\usepackage[breaklinks,colorlinks,allcolors=cvprblue]{hyperref}
\renewcommand*{\HyperDestNameFilter}[1]{\jobname-#1}
\hypersetup{
  pdftitle={MoVelo: World-Grounded Human Motion Recovery with Feedback from Conditional Velocity Field},
  pdfauthor={Da Li, Xunshu Yang, Jiping Jin, Ming Li, Ce Zheng}
}
\makeatletter
\renewcommand{\paragraph}{\@startsection{paragraph}{4}{\z@}%
  {6pt plus 2pt minus 1pt}{-0.8em}{\normalfont\normalsize\bfseries}}
\makeatother

\makeatletter
\patchcmd{\@maketitle}{\vskip .375in}{\vskip 6pt}{}{}
\patchcmd{\@maketitle}{\vspace*{24pt}}{\vspace*{15pt}}{}{}
\patchcmd{\@maketitle}{\vspace*{12pt}}{\vspace*{7pt}}{}{}
\g@addto@macro{\@maketitle}{%
\par\noindent
\begin{minipage}{\textwidth}
\captionsetup{hypcap=false}
\centering
\makebox[\linewidth][c]{\includegraphics[width=0.92\linewidth]{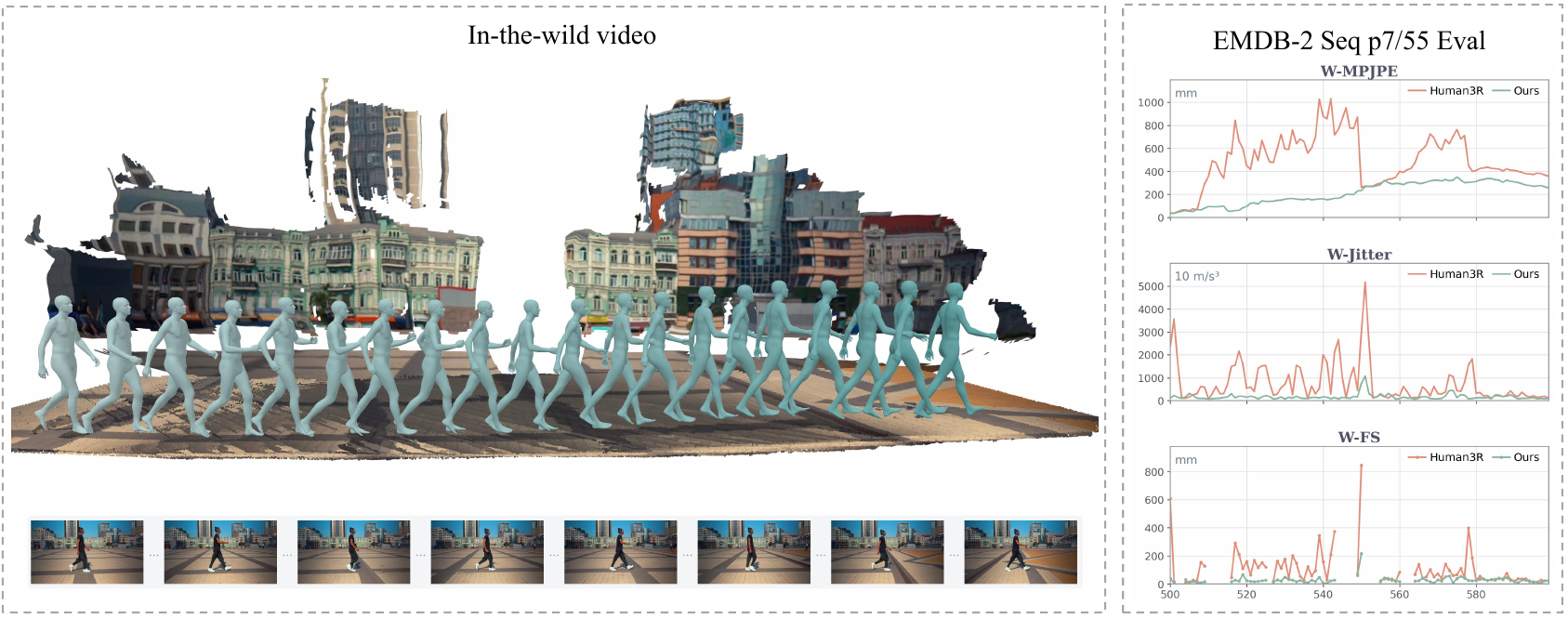}}
\captionof{figure}{Overview of MoVelo on a walking sequence. Reconstructed human
meshes and scene geometry are shown above the input frames; the right panels
compare world-space MPJPE, jitter, and foot sliding (W-FS) with Human3R
on EMDB-2~\citep{kaufmann2023emdb}.}
\label{fig:teaser}
\end{minipage}\par
\vspace{\textfloatsep}%
}
\makeatother

\begin{document}
\flushbottom
\setlength{\floatsep}{7pt plus 2pt minus 1pt}
\setlength{\textfloatsep}{10pt plus 2pt minus 2pt}
\setlength{\dblfloatsep}{7pt plus 2pt minus 1pt}
\setlength{\dbltextfloatsep}{10pt plus 2pt minus 2pt}
\setlength{\intextsep}{8pt plus 2pt minus 1pt}
\setlength{\abovedisplayskip}{7pt plus 2pt minus 2pt}
\setlength{\belowdisplayskip}{7pt plus 2pt minus 2pt}
\clubpenalty=3000
\widowpenalty=3000
\displaywidowpenalty=3000
\setcounter{footnote}{1}
\maketitle
\begingroup
\brokenpenalty=10000
\begin{abstract}
\par
Online human motion reconstruction aims to recover 3D body motion in a common world coordinate system from a monocular video stream, using only current and past frames. Occlusions and ambiguity between body and camera motion make temporal coherence particularly challenging, as inconsistent frame-to-frame estimates can produce jitter and unreliable global trajectories. We introduce MoVelo, which uses explicit motion feedback to refine latent human representations for current-frame mesh prediction. To encourage temporal consistency while allowing corrections from new observations, MoVelo explicitly links successive body states through a conditional velocity field. Conditioned on temporal tokens and previously reconstructed body states, this field relates joint positions, velocities, and accelerations in a fixed world reference through integration and differentiation. Inspired by physics-informed neural networks, we jointly supervise these quantities and feed field-derived motion features back into the human representations before mesh regression. Experiments on EMDB-2 and RICH demonstrate improvements in global motion accuracy and temporal consistency. Compared with Human3R, MoVelo reduces W-MPJPE from 267.9 to 257.9 mm on EMDB-2 and from 184.9 to 170.7 mm on RICH. It also reduces world-coordinate jitter by 9.9\% and 14.5\%, respectively, with lower foot sliding on both datasets. These results highlight the value of explicit motion feedback for temporally coherent online human motion reconstruction in world coordinates. Our implementation is available on the project page: {\urlstyle{same}\url{https://dali936.github.io/Movelo-page/}}.
\end{abstract}
\par
\endgroup
\section{Introduction}
\label{sec:introduction}

Online human motion reconstruction
aims to recover temporally
coherent 3D body motion in a shared world coordinate system as
a monocular video stream unfolds.
Recovering both
articulated body poses and global trajectories supports applications in augmented
reality and humanoid whole-body
control.
This task is challenging: image motion combines body articulation,
human translation, and camera movement, while occlusions leave parts of the
body unobserved. An online method must interpret this incomplete evidence
using only the current and preceding frames~\citep{zhao2026onlinehmr}. Uncertainty in these estimates
can vary across frames, producing jitter and unreliable global trajectories.

Maintaining coherent motion therefore requires reasoning about how the
reconstructed body changes over time.
Human3R~\citep{chen2026human3r} provides a unified online framework by
combining human visual prompts with CUT3R's recurrent scene
representation~\citep{wang2025cut3r}. Its recurrent state carries evidence
across frames, but the relation between successive body states remains
implicit: \textit{the representation does not specify the joint velocities and
accelerations connecting them.}
We seek an
explicit link between successive body states that guides current-frame
reconstruction while allowing new observations to correct the motion estimate.

We therefore introduce \textbf{MoVelo}, a method for
online human motion reconstruction that combines a conditional velocity
field with motion feature feedback. For each tracked person, the field
predicts velocities from a joint configuration, a temporal token, and
time. The token gathers current visual evidence and recurrent context,
while joints from the preceding reconstructed mesh anchor the transition.
Integrating the field gives an updated joint state; differentiating the
same field along its predicted motion gives acceleration. Inspired by
physics-informed learning~\citep{raissi2019pinn}, we jointly supervise
these linked positions, velocities, and accelerations.
Motion feature
feedback uses the predicted velocities and integrated displacements,
together with the temporal token, to refine human representations before
SMPL-X regression.
This path lets the current mesh prediction use the estimated motion,
while mesh reconstruction losses train the field through the same path.
Joints from the final mesh are transformed into the fixed world
reference using the predicted camera pose, then stored as the next motion state.

Experiments show that MoVelo improves global joint accuracy over
Human3R on EMDB-2~\citep{kaufmann2023emdb} and RICH~\citep{huang2022capturing}.
MoVelo also reduces world-coordinate
jitter, acceleration error, and foot sliding on both datasets.
Ablations show the benefits of the
conditional velocity field and motion feature feedback, and examine how
joint-position and acceleration supervision affect the field.
Figure~\ref{fig:teaser} visualizes the recovered human motion together
with scene geometry in a shared world coordinate system.

Our contributions are:
\nopagebreak
\begingroup
\settowidth{\leftmargini}{\labelitemi}
\addtolength{\leftmargini}{\labelsep}
\begin{itemize}
\item We formulate a conditional velocity field for online human motion
reconstruction, explicitly representing body-state transitions
in a shared world coordinate system and linking
joint positions, velocities, and accelerations through
integration, differentiation, and combined supervision.
\item We introduce motion feature feedback that encodes predicted velocities,
integrated displacements, and decoded temporal tokens to refine human
representations before mesh regression, enabling current-frame reconstruction
to use the estimated body-state transition.
\item We demonstrate improved global motion accuracy and temporal
consistency on EMDB-2 and RICH, and use progressive ablations to assess
the contributions of velocity field supervision and motion feature feedback.
\end{itemize}
\endgroup
\section{Related Work}
\label{sec:related_work}

\paragraph{Human Mesh and Motion Recovery.}
Parametric body models provide a common representation for human
mesh recovery.
SMPL~\citep{loper2015smpl} represents body shape and pose,
SMPL-X~\citep{pavlakos2019smplx} adds articulated hands and facial
expressions, and SKEL~\citep{keller2023skel} combines the body surface
with a biomechanical skeleton.
HMR~\citep{kanazawa2018hmr}, SPIN~\citep{kolotouros2019spin}, and
HMR~2.0~\citep{goel2023humans4d} estimate SMPL parameters from images using direct
regression, fitting-based supervision, and transformer architectures,
respectively.
Multi-HMR~\citep{baradel2024multihmr} extends image-based recovery to
multiple SMPL-X bodies, while HSMR~\citep{xia2025hsmr} and
SKEL-CF~\citep{li2026skelcf} recover SKEL parameters.
Temporal methods reduce frame-level ambiguity by aggregating information across
time. VIBE~\citep{kocabas2020vibe} and TCMR~\citep{choi2021tcmr} model video
features, while TCMR uses bidirectional context from both past and future
frames.
Recovering global human motion from monocular video requires modeling the
coupled effects of human and camera motion. GLAMR~\citep{yuan2022glamr} and
SLAHMR~\citep{ye2023slahmr} use motion priors and trajectory optimization to
estimate global trajectories. WHAM~\citep{shin2024wham} incorporates camera
motion and contact-aware trajectory refinement, while TRAM~\citep{wang2024tram}
combines metric-scale camera estimation with local body motion. GVHMR~\citep{shen2024world}
uses gravity-view coordinates to recover world-grounded motion.

\paragraph{Joint Human--Scene Reconstruction.}
Scene geometry provides context for human scale, placement, and contact,
motivating joint estimation of the body and its surroundings.
PhySIC~\citep{ym2025physic} uses depth and contact estimates to optimize
human--scene configurations from a single image.
JOSH~\citep{liu2026josh} jointly optimizes human motion, camera
trajectories, and dense geometry from monocular video using contact
constraints. Other acquisition settings supply richer observations: two
moving RGB-D cameras in EmbodMocap~\citep{wang2026embodmocap} and multiple
views in CHROMM~\citep{kim2026chromm}.
Feed-forward methods learn the alignment between human and scene
predictions. JOSH3R~\citep{liu2026josh} learns pairwise human transformations from
MASt3R~\citep{leroy2024mast3r} features using supervision from JOSH's
optimized reconstructions.
UniSH~\citep{li2026unish} combines scale and body-placement prediction
with geometry distillation, while SHOW~\citep{shi2026show} uses mask prompts
and geometry-aware SMPL-X decoding.
For online reconstruction, human and scene estimates need to update as
frames arrive.
OnlineHMR~\citep{zhao2026onlinehmr} combines a causal human estimation
branch with incremental SLAM and regularizes joint velocities and
accelerations during training.
Human3R~\citep{chen2026human3r} uses visual prompts to combine
CUT3R's recurrent scene state~\citep{wang2025cut3r} with
Multi-HMR's human priors~\citep{baradel2024multihmr}.
UniCon3R~\citep{sur2026unicon3r} uses geometry and contact history to
update human latents before mesh regression.
\paragraph{Physics-Informed Motion Modeling.}
Prior work uses motion priors and physical constraints to encourage
temporal coherence and physical plausibility in human reconstruction.
HuMoR~\citep{rempe2021humor} learns body-state transitions as a prior for
motion fitting, while HTD-Refine~\citep{wei2026htdrefine} predicts velocities
and accelerations to guide sequence optimization.
QuaMo~\citep{le2026quamo} formulates motion through quaternion differential
equations and a meta-PD controller with acceleration enhancement, using
integration to update the reconstructed motion.
PhysPT~\citep{zhang2024physpt} incorporates differentiable body dynamics,
contact forces, and Euler--Lagrange constraints into motion reconstruction.
Simulation-based approaches include PhysHMR~\citep{feng2025physhmr}, which
learns a humanoid control policy from visual features through distillation
and reinforcement learning, and RePHO~\citep{huang2026repho}, which refines
human--object interactions through simulated tracking and updates to
reference trajectories.
Physics-informed neural networks couple observations with governing
equations, using automatic differentiation to evaluate their
residuals~\citep{raissi2019pinn,luo2025pinnreview}.
Drawing on this principle, we formulate a conditional velocity field whose
integration and differentiation couple joint positions, velocities, and
accelerations in a fixed world reference.
We supervise these quantities jointly and feed the estimated
motion back into human representations to guide online body reconstruction.
\section{Method}
\label{sec:method}

Given a monocular video $\{I_t\}_{t=1}^{T}$, MoVelo reconstructs human
meshes, camera poses, and scene geometry online from observations available
up to frame $t$. Figure~2 summarizes the pipeline. For each tracked person,
a shared recurrent decoder outputs human and temporal tokens. The temporal
token conditions a velocity field that integrates the joint state
in a fixed world reference from the
preceding reconstruction to estimate the current state. The decoded temporal
token, predicted velocities, and integrated displacements form motion features
that refine the human representation before SMPL-X regression. The resulting
mesh supplies the joint state after transformation into this
world reference using the predicted camera pose for the next transition.

\begin{figure*}[tp]
\centering
\includegraphics[width=0.94\linewidth]{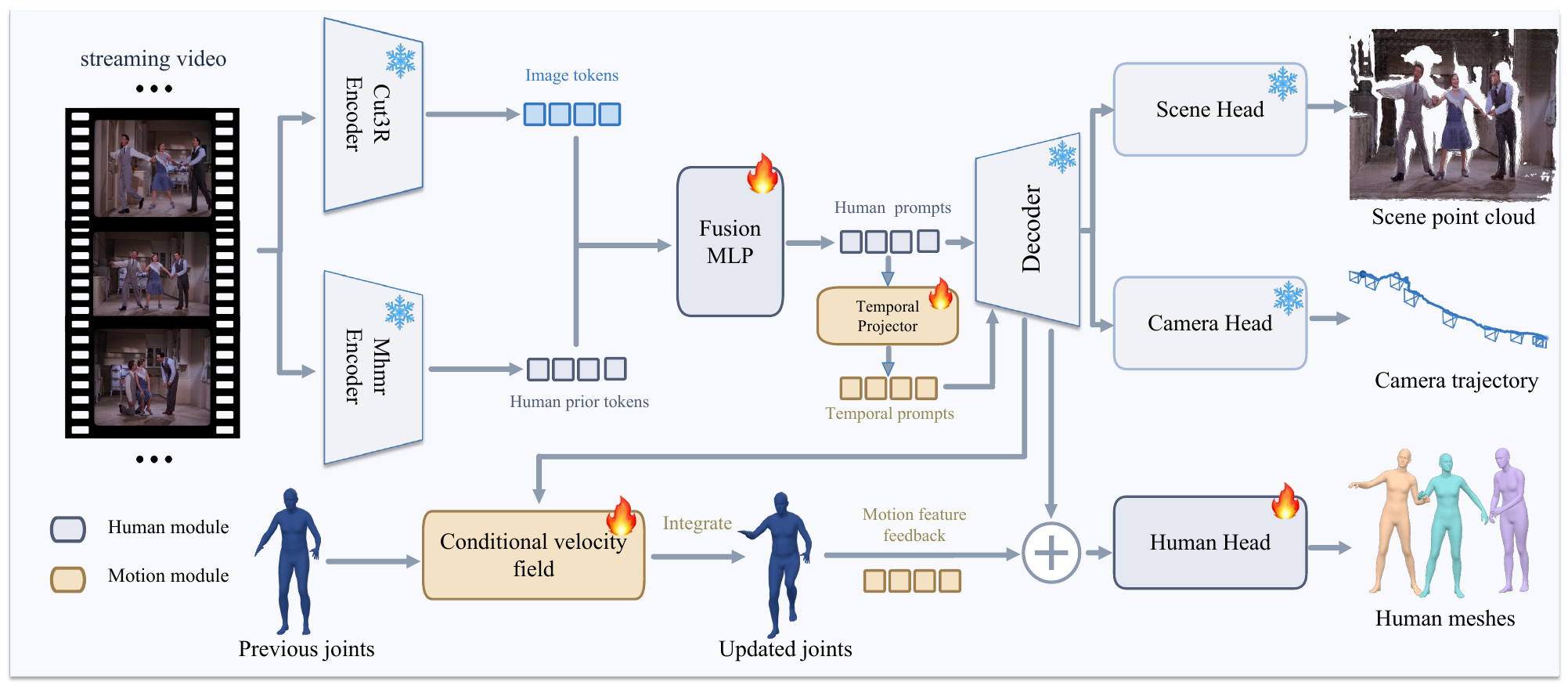}
\caption{\textbf{Overview of MoVelo.} Frozen CUT3R and Multi-HMR features
generate human prompts and detached temporal prompts. Within a shared frozen
decoder, the human and temporal prompts interact with image tokens and
recurrent scene memory. The decoder's temporal outputs condition a
conditional velocity field for joint integration in a fixed world
reference. The temporal tokens and the
field outputs, including predicted velocities and integrated displacements,
jointly form motion feedback that refines the human latents. The human head
then uses the resulting refined features for SMPL-X regression.
The reconstructed joints are transformed into the world reference
using the predicted camera pose and stored for the next transition.}
\label{fig:framework}
\end{figure*}

\subsection{Preliminaries}
\label{sec:method-representation}

Following Human3R~\citep{chen2026human3r}, we fuse CUT3R scene
features~\citep{wang2025cut3r} and Multi-HMR human-prior
features~\citep{baradel2024multihmr} with an MLP to form a human prompt
$\mathbf{q}_t^{\mathrm{human}}$ for each detected person.
A temporal prompt is initialized as
\begin{equation*}
\mathbf{q}_t^{\mathrm{temp}}
=\operatorname{MLP}\!\bigl(\operatorname{sg}(\mathbf{q}_t^{\mathrm{human}})\bigr),
\end{equation*}
where $\operatorname{sg}$ denotes stop-gradient.
Both prompts share the same image location and interact with image tokens
and persistent scene memory in the recurrent decoder, producing the human
token $\mathbf{h}_t$ and temporal token $\mathbf{z}_t$.
We denote the associated human-prior feature by $\mathbf{p}_t$.
\par
\noindent
For each tracked person, let
$\mathbf{X}_t\in\mathbb{R}^{J\times3}$ denote the
world-space
positions of the SMPL-X body joints ($J=22$) in the reconstructed body at
frame $t$.
We anchor the world reference to the first camera frame of each
sequence, or sampled training clip, and keep it fixed thereafter.
Let $\mathcal{T}_t$ denote the predicted camera-to-world rigid transform
at frame $t$, applied to each joint as
$\mathcal{T}_t(\mathbf{x})=\mathbf{R}_t\mathbf{x}+\mathbf{t}_t$.
We obtain $\mathbf{X}_t$ by applying this transform to the final-mesh joints.
The sequence $\{\mathbf{X}_t\}_{t=1}^{T}$ records the person's joint states
over the video. For online inference, the motion state at frame $t$ is
initialized with a detached copy of the final-mesh joints reconstructed at
the previous time step,
$\mathbf{X}^{\mathrm{history}}_{t-1}=\operatorname{sg}(\mathbf{X}_{t-1})$.
If $\tau_t$ denotes the timestamp of frame $t$, then a valid transition has
$\Delta_t=\tau_t-\tau_{t-1}>0$, measured in seconds.
All joint states are measured in meters in this common world
reference, so temporal differences use the same coordinate frame.
\begin{figure}[tbp]
\centering
\includegraphics[width=\linewidth]{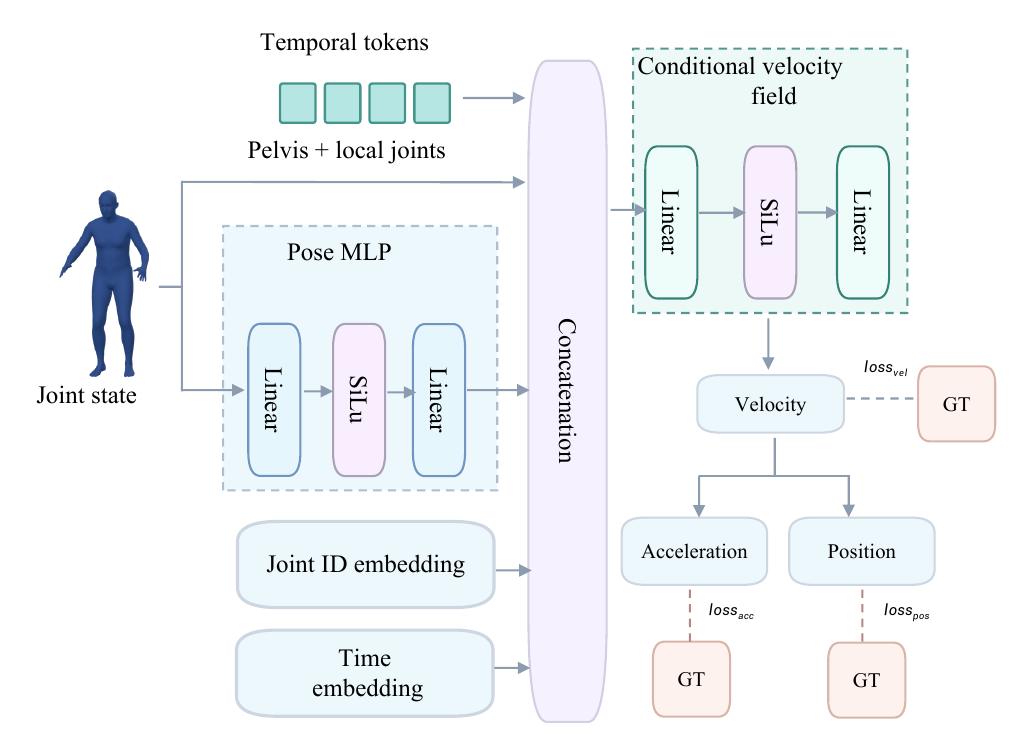}
\caption{\textbf{Conditional velocity field.} Given a world-space
joint state and temporal features, the field predicts per-joint
world-space velocities.
We integrate the velocity field to obtain joint positions and differentiate
it along the joint trajectory to obtain accelerations.}
\label{fig:velocity-field}
\end{figure}
\par

\subsection{Conditional Velocity Field}
\label{sec:method-field}

\paragraph{State and visual conditioning.}
We model each body-state transition with a conditional velocity field
$f_\phi(\mathbf{X},\mathbf{z}_t,s,\Delta_t)\in\mathbb{R}^{J\times3}$.
Fig.~\ref{fig:velocity-field} summarizes its architecture and supervision.
For joint $j$, the network predicts
\begin{equation}
 \setlength{\jot}{3pt}
 \begin{aligned}
 \widetilde{\mathbf{v}}_j = F_\phi\!\Bigl(\mathbf{z}_t
 &{}\oplus \mathbf{e}_j \oplus \mathbf{x}_0
   \oplus (\mathbf{x}_j-\mathbf{x}_0)\\
 &{}\oplus E_{\mathrm{pose}}\!\bigl(
   \operatorname{Flatten}(\mathbf{X}-\mathbf{x}_0)\bigr)\\
 &{}\oplus E_s(s) \oplus E_\Delta(\Delta_t)\Bigr),
 \end{aligned}
 \label{eq:field-input}
\end{equation}
\noindent
where $\oplus$ denotes concatenation along the channel axis,
and $\mathbf{z}_t$ is the recurrent decoder's output for the temporal
prompt, combining current visual evidence with recurrent context.
The learned embedding $\mathbf{e}_j$ identifies joint $j$.
The pelvis position $\mathbf{x}_0$ retains the body's
world-space
location, and $\mathbf{x}_j-\mathbf{x}_0$ gives joint $j$'s position
relative to the pelvis.
These relative positions also use the fixed world axes.
$E_{\mathrm{pose}}$ encodes the full pelvis-relative joint configuration
$\mathbf{X}-\mathbf{x}_0$, where $\mathbf{X}$ contains the $J$ joint
positions input to the velocity field.
The operator $\operatorname{Flatten}$ reshapes this $J\times3$ matrix
into a $3J$-dimensional vector. The resulting whole-body feature couples
each joint's velocity to the overall configuration.
$E_s$ and $E_\Delta$ separately embed the query time $s$ and the elapsed
time between frames $\Delta_t$, both in seconds, with $s\in[0,\Delta_t]$.
These three embedding functions are learned MLPs. The field predicts
standardized velocities $\widetilde{\mathbf{v}}$,
which we convert to physical velocities in meters per second as
$f_\phi=\boldsymbol{\mu}_v+\boldsymbol{\sigma}_v\odot\widetilde{\mathbf{v}}$.
We compute $\boldsymbol{\mu}_v$ and $\boldsymbol{\sigma}_v$ as the mean
and standard deviation of ground-truth velocities for each joint and
coordinate. These velocities are obtained by finite differences of
annotated world-space joint positions
expressed in the fixed clip reference in the training data; the
statistics remain fixed during training and inference.

\Needspace{3\baselineskip}
\paragraph{Integrating the body state.}
For each valid transition, we initialize the joint state
$\mathbf{X}_0=\mathbf{X}^{\mathrm{history}}_{t-1}$ with the
detached world-space joint positions reconstructed in the preceding frame.
Let $\mathbf{X}(s)$ denote the joint positions obtained by integrating
the conditional velocity field from $\mathbf{X}_0$ for an elapsed time
$s$, where $s$ ranges from $0$ to the frame interval $\Delta_t$.
The evolution of these positions is governed by the following
differential equation and initial condition:
\begin{equation}
 \begin{aligned}
 \frac{d\mathbf{X}(s)}{ds}
 &= f_\phi\!\bigl(\mathbf{X}(s),\mathbf{z}_t,s,\Delta_t\bigr),\\[2pt]
 \mathbf{X}(0)&=\mathbf{X}_0.
 \end{aligned}
 \label{eq:motion-ode}
\end{equation}
The conditional velocity field $f_\phi$ predicts the instantaneous
joint velocities at the intermediate joint positions $\mathbf{X}(s)$.
During integration, the conditioning
token $\mathbf{z}_t$ and the total interval $\Delta_t$ remain fixed,
while the joint positions and elapsed time vary.
To estimate the joint positions at the current frame, we integrate the
conditional velocity field over the frame interval $\Delta_t$ using a
single step of Heun's method~\mbox{\citep{hairer1993solving}}:
\begin{equation}
 \setlength{\jot}{3pt}
 \begin{alignedat}{2}
 &\vphantom{\mathbf{X}^{\mathrm{field}}_t}\mathbf{v}_0
 &&= f_\phi\!\bigl(\mathbf{X}_0,\mathbf{z}_t,0,\Delta_t\bigr),\\
 &\vphantom{\mathbf{X}^{\mathrm{field}}_t}\mathbf{X}_{\mathrm{E}}
 &&= \mathbf{X}_0 + \Delta_t\,\mathbf{v}_0,\\
 &\vphantom{\mathbf{X}^{\mathrm{field}}_t}\mathbf{v}_1
 &&= f_\phi\!\bigl(\mathbf{X}_{\mathrm{E}},\mathbf{z}_t,\Delta_t,\Delta_t\bigr),\\
 &\vphantom{\mathbf{X}^{\mathrm{field}}_t}\mathbf{v}_t
 &&= \tfrac{1}{2}\bigl(\mathbf{v}_0 + \mathbf{v}_1\bigr),\\
 &\mathbf{X}^{\mathrm{field}}_t\;
 &&= \mathbf{X}_0 + \Delta_t\,\mathbf{v}_t.
 \end{alignedat}
 \label{eq:heun}
\end{equation}
\noindent
The same velocity-field network predicts $\mathbf{v}_0$ at the initial
joint positions and $\mathbf{v}_1$ at the provisional joint positions
$\mathbf{X}_{\mathrm{E}}$.
Multiplying their average velocity $\mathbf{v}_t$ by the frame interval
gives the joint displacement. Adding this displacement to the initial joint positions
$\mathbf{X}_0$ yields the estimated current-frame joint positions
$\mathbf{X}^{\mathrm{field}}_t$.
We supervise these positions using ground-truth joints and encode the
displacement as part of the motion features for feedback.

\Needspace{3\baselineskip}
\paragraph{Deriving acceleration from the velocity field.}
To compute joint accelerations by differentiating the velocity field,
we use SiLU activations in the velocity-field network and its embedding
MLPs to ensure smooth derivatives with respect to joint positions and time.
Along the joint trajectory in Eq.~\ref{eq:motion-ode}, acceleration is
the rate of change of the predicted velocity:
\begin{equation}
 \begin{aligned}
 \mathbf{v}(s)&=f_\phi\!\bigl(\mathbf{X}(s),\mathbf{z}_t,s,\Delta_t\bigr),\\[2pt]
 \mathbf{a}(s)&=\frac{d\mathbf{v}(s)}{ds}.
 \end{aligned}
 \label{eq:field-acceleration}
\end{equation}
Here $\mathbf{v}(s)$ is the instantaneous joint velocity and
$\mathbf{a}(s)$ is the corresponding acceleration.
Both derivatives are taken in the fixed world reference.
The total derivative accounts for both the direct effect of time on
the predicted velocities and the changes in joint positions along the
trajectory, while $\mathbf{z}_t$ and $\Delta_t$ remain fixed.
We compute the current-frame acceleration $\mathbf{a}^{\mathrm{field}}_t$
at $s=\Delta_t$, using the estimated joint positions
$\mathbf{X}^{\mathrm{field}}_t$ from Eq.~\ref{eq:heun}.
We compute these accelerations using PyTorch's automatic differentiation.

\subsection{Motion Feature Feedback}
\label{sec:method-feedback}

The field estimates a joint-state transition, while the body
regressor produces the final parametric mesh. Feature feedback connects
these predictions: it allows the estimated transition to refine body
features and allows mesh losses to supervise the field through this path.
We use an MLP to encode the two standardized world-space
velocity estimates, the world-space joint displacement,
and an MLP embedding of the frame interval.
Joint arrays are flattened before concatenation.
The resulting motion feature $\mathbf{r}_t$ updates the temporal token
$\mathbf{z}_t$ as follows:
\begin{equation}
 \begin{aligned}
 \mathbf{r}_t &= \operatorname{MLP}\!\left(
 \begin{aligned}
 &\widetilde{\mathbf{v}}_0 \oplus \widetilde{\mathbf{v}}_1
   \oplus (\mathbf{X}^{\mathrm{field}}_t-\mathbf{X}_0)\\
 &\quad\oplus \operatorname{MLP}(\Delta_t)
 \end{aligned}
 \right),\\[3pt]
 \mathbf{z}'_t &= \mathbf{z}_t + m_t\tanh(\mathbf{r}_t).
 \end{aligned}
 \label{eq:motion-feedback}
\end{equation}
Here $m_t$ is a binary mask that gates the field-derived increment to the temporal token.
It equals $1$ when the current person has a valid identity match in the
preceding frame and $\Delta_t>0$, and $0$ otherwise.
Another MLP takes the updated temporal token and the human-prior feature
as input and predicts additive updates to the human token and human-prior
feature:
\begin{equation}
 \mathbf{h}'_t \oplus \mathbf{p}'_t
 = (\mathbf{h}_t \oplus \mathbf{p}_t)
 + \operatorname{MLP}(\mathbf{z}'_t \oplus \mathbf{p}_t).
 \label{eq:human-feedback}
\end{equation}
The human head regresses SMPL-X~\citep{pavlakos2019smplx} pose, shape,
and expression from $\mathbf{h}'_t \oplus \mathbf{p}'_t$, and camera-space
translation from $\mathbf{h}'_t$.
Let $\mathcal{J}$ extract the first 22 body joints from the decoded mesh.
We update the motion history as
\begin{equation}
 \mathbf{X}^{\mathrm{history}}_t
 = \operatorname{sg}\!\left(
 \mathcal{T}_t\!\left(\mathcal{J}(\mathcal{M}_t)\right)\right),
 \label{eq:motion-history}
\end{equation}
where $\mathcal{M}_t$ is the final camera-space mesh.
We store the reconstructed world-space joints as the starting positions for the next
frame and treat them as fixed inputs.
The stop-gradient also blocks back-propagation through the camera
transform used to construct this history.
Within each frame, the human
reconstruction loss also trains the velocity field through motion feature
feedback.

\Needspace{3\baselineskip}
\subsection{Training Strategy}
\label{sec:method-training}

Following the unified online method Human3R~\citep{chen2026human3r},
we train MoVelo on the synthetic BEDLAM dataset~\citep{black2023bedlam}
using scene, camera, and SMPL-X annotations. We retain RGB reconstruction,
confidence-aware pointmap regression, and camera-pose supervision in
$\mathcal{L}_{\mathrm{recon}}$. Human reconstruction is supervised by
$\mathcal{L}_{\mathrm{human}}$, which applies L1 losses to SMPL-X
parameters, 3D joints and vertices, and their image reprojections.
Our temporal loss $\mathcal{L}_{\mathrm{temporal}}$ supervises the
velocity field through integrated joint positions, interval-average
velocities, and endpoint accelerations.
The joint-position, velocity, and acceleration targets are all
expressed in the fixed world reference of the training clip.
We jointly train the
human prompt and reconstruction head together with the temporal prompt,
velocity field, and feedback networks, while keeping the pretrained
scene and human-prior modules frozen. The full objective, supervision
targets, loss weights, and optimization details are provided in
Section~A of Sup.Mat.
\section{Experiments}
\label{sec:experiments}

We evaluate MoVelo in terms of global human motion accuracy and temporal
consistency. Quantitative comparisons measure
reconstruction accuracy and motion quality, while qualitative results
visualize the recovered humans and scenes in a shared coordinate system.
Controlled ablations further examine the contributions of the conditional
velocity field and motion feature feedback. Additional field-supervision
analyses and local mesh reconstruction results are provided in
Section~C of Sup.Mat.

\subsection{Datasets and Evaluation Metrics}
\label{sec:exp-setup}

We evaluate global motion and temporal consistency on EMDB-2
(25 sequences; \citealp{kaufmann2023emdb}) and RICH
(191 static-camera sequences; \citealp{huang2022capturing}). We report
W-MPJPE, WA-MPJPE, and RTE for global motion, and W-Jitter (predicted jerk),
W-Accel (acceleration error), and W-FS (foot sliding) for temporal
consistency in world coordinates.
Lower is better for all metrics. Further evaluation details are
provided in
Section~A of Sup.Mat.

\begin{table*}[tp]
\centering
\caption{Global human motion on EMDB-2 and RICH.
\textbf{Unified}: joint human, scene, and camera prediction within one model;
\textbf{Online}: causal world-space inference.}
\label{tab:global-motion}
\papertablestyle
\setlength{\tabcolsep}{2.5pt}
\begin{tabularx}{\textwidth}{lcc*{2}{
>{\hsize=1.20\hsize\linewidth=\hsize\centering\arraybackslash}X
>{\hsize=1.10\hsize\linewidth=\hsize\centering\arraybackslash}X
>{\hsize=0.70\hsize\linewidth=\hsize\centering\arraybackslash}X}}
\toprule
\multirow{2}{*}{Method}
& \multirow{2}{*}{Unified}
& \multirow{2}{*}{Online}
& \multicolumn{3}{c}{EMDB-2~\citep{kaufmann2023emdb}}
& \multicolumn{3}{c}{RICH~\citep{huang2022capturing}} \\
\cmidrule(lr){4-6}\cmidrule(lr){7-9}
& &
& \multicolumn{1}{c}{WA-MPJPE $\downarrow$}
& \multicolumn{1}{c}{W-MPJPE $\downarrow$}
& \multicolumn{1}{c}{RTE $\downarrow$}
& \multicolumn{1}{c}{WA-MPJPE $\downarrow$}
& \multicolumn{1}{c}{W-MPJPE $\downarrow$}
& \multicolumn{1}{c}{RTE $\downarrow$} \\
\midrule
GVHMR~\citep{shen2024world} & \ding{55} & \ding{55} & 111.0 & 276.5 & 2.0 & \textbf{78.8} & \textbf{126.3} & \underline{2.4} \\
TRAM~\citep{wang2024tram} & \ding{55} & \ding{55} & \underline{76.4} & 222.4 & \underline{1.4} & 127.8 & 238.0 & 6.0 \\
JOSH$_3$~\citep{liu2026josh} & \ding{55} & \ding{55} & \textbf{68.9} & \textbf{174.7} & \textbf{1.3} & -- & -- & -- \\
UniSH~\citep{li2026unish} & \ding{51} & \ding{55} & 118.5 & 270.1 & 5.8 & 118.1 & 183.2 & 4.8 \\
SHOW~\citep{shi2026show} & \ding{51} & \ding{55} & 109.1 & 262.3 & 2.1 & 107.3 & 172.7 & \textbf{2.2} \\
CHROMM~\citep{kim2026chromm} & \ding{51} & \ding{55} & 102.6 & 255.0 & 1.7 & \underline{87.5} & \underline{138.3} & 3.3 \\
EmbodMocap~\citep{wang2026embodmocap} & \ding{55} & \ding{55} & 82.2 & \underline{220.7} & 1.7 & -- & -- & -- \\
WHAM~\citep{shin2024wham} & \ding{55} & \ding{55} & 135.6 & 354.8 & 6.0 & 108.4 & 196.1 & 4.5 \\
\midrule
JOSH3R~\citep{liu2026josh} & \ding{55} & \ding{51} & 220.0 & 661.7 & 13.1 & -- & -- & -- \\
OnlineHMR~\citep{zhao2026onlinehmr} & \ding{55} & \ding{51} & \textbf{93.5} & 310.4 & \underline{2.2} & -- & -- & -- \\
UniCon3R~\citep{sur2026unicon3r} & \ding{51} & \ding{51} & 113.7 & 285.6 & 2.3 & -- & -- & -- \\
Human3R~\citep{chen2026human3r} & \ding{51} & \ding{51} & 112.2 & \underline{267.9} & \underline{2.2} & \underline{110.0} & \underline{184.9} & \underline{3.3} \\
\rowcolor{tablehighlight}
\textbf{MoVelo} & \ding{51} & \ding{51} & \underline{107.7} & \textbf{257.9} & \textbf{2.1} & \textbf{105.1} & \textbf{170.7} & \textbf{3.1} \\
\bottomrule
\end{tabularx}
\end{table*}
\begin{table}[tbp]
\centering
\caption{Inference efficiency and EMDB-2 W-MPJPE$_{100}$.
Delay is in seconds; error is in mm. The two MoVelo variants use ViT-S
(Lightweight) and ViT-L/896 (Final) human-prior encoders.}
\label{tab:efficiency}
\papertablestyle
\renewcommand{\arraystretch}{1.16}
\setlength{\tabcolsep}{2pt}
\renewcommand{\tabularxcolumn}[1]{m{#1}}
\begin{tabularx}{\columnwidth}{>{\raggedright\arraybackslash}Xccc}
\toprule
Method & FPS $\uparrow$ & Avg. delay $\downarrow$ & W-MPJPE$_{100}$ $\downarrow$ \\
\midrule
\multicolumn{4}{l}{\textit{Offline}} \\
GLAMR~\citep{yuan2022glamr} & 17.2 & 14.16 & 726.6 \\
SLAHMR~\citep{ye2023slahmr} & 0.1 & 2435.00 & 776.1 \\
WHAM (w/ DPVO)~\citep{shin2024wham} & 9.3 & 26.18 & 354.8 \\
TRAM~\citep{wang2024tram} & 2.1 & 115.95 & 222.4 \\
GVHMR (w/ DPVO)~\citep{shen2024world} & 12.1 & 20.12 & 276.5 \\
\midrule
\multicolumn{4}{l}{\textit{Online}} \\
TRACE~\citep{sun2023trace} & 24.6 & 0.04 & 1702.3 \\
Human3R~\citep{chen2026human3r} & 4.8 & 0.21 & 267.9 \\
OnlineHMR~\citep{zhao2026onlinehmr} & 3.3 & 0.30 & 310.4 \\
\rowcolor{tablehighlight}
\textbf{MoVelo-Lightweight} & 10.8 & 0.09 & -- \\
\rowcolor{tablehighlight}
\textbf{MoVelo-Final} & 4.6 & 0.22 & 257.9 \\
\bottomrule
\end{tabularx}
\end{table}
\subsection{Comparison with Existing Methods}
\label{sec:exp-reconstruction}

\paragraph{Global human motion.}
MoVelo achieves W-MPJPEs of 257.9\,mm on EMDB-2 and 170.7\,mm on RICH,
outperforming Human3R by 3.7\% and 7.7\%, respectively
(Table~\ref{tab:global-motion}).
WA-MPJPE decreases to 107.7 and 105.1\,mm, while RTE reaches
2.1\% and 3.1\%, respectively, improving over Human3R on both datasets.
Among online methods, MoVelo achieves the lowest RTE on both benchmarks
and leads across all three metrics on RICH, demonstrating
accurate world-space joint localization and trajectory recovery.

\paragraph{Inference efficiency.}
\label{sec:exp-efficiency}
Table~\ref{tab:efficiency} highlights MoVelo's favorable balance between
online reconstruction accuracy and inference cost. The lightweight variant
reaches 10.8 FPS with 0.09\,s average delay. The final model achieves the
lowest reported W-MPJPE$_{100}$ among the compared online methods
(257.9\,mm), while retaining 4.6 FPS and 0.22\,s delay. Compared with
Human3R, it reduces error by 3.7\% with only 0.01\,s additional reported
average delay. Both variants use only current and preceding frames.
The final model also improves on OnlineHMR in both accuracy and throughput
(4.6 vs.\ 3.3 FPS). The lightweight variant reduces average delay to less
than half that of the final model, offering an option for applications
that prioritize responsiveness.
Training is also economical: we optimize only 73.23M parameters
(6.24\% of the model) in approximately one day on a single RTX 4090.

\begingroup
\begin{figure*}[tp]
\centering
\includegraphics[width=\linewidth]{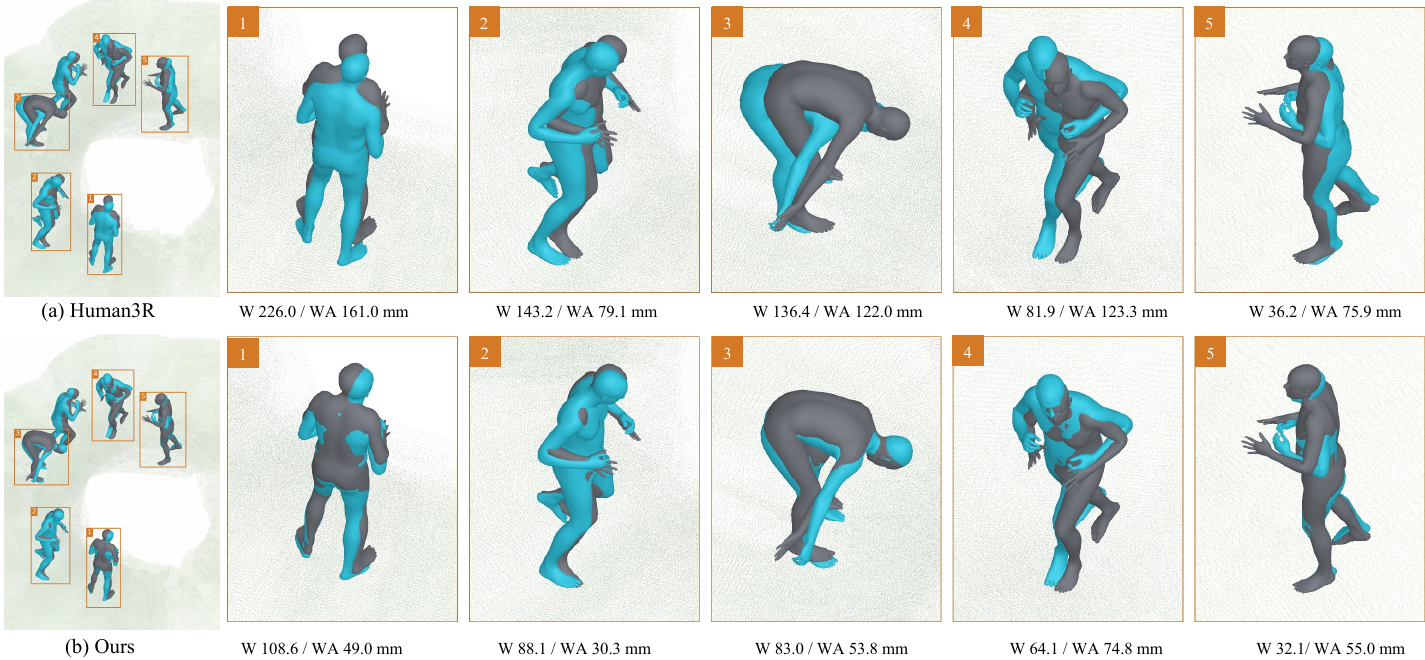}
\caption{\textbf{World-space reconstruction on EMDB-2.}
Human3R (top) and MoVelo (bottom) predictions are shown in cyan against
gray ground truth. We report the W (W-MPJPE) and WA (WA-MPJPE) errors for each of the five examples.}
\label{fig:world-reconstruction-emdb2}
\end{figure*}

\begin{figure*}[tp]
\centering
\includegraphics[width=0.90\linewidth]{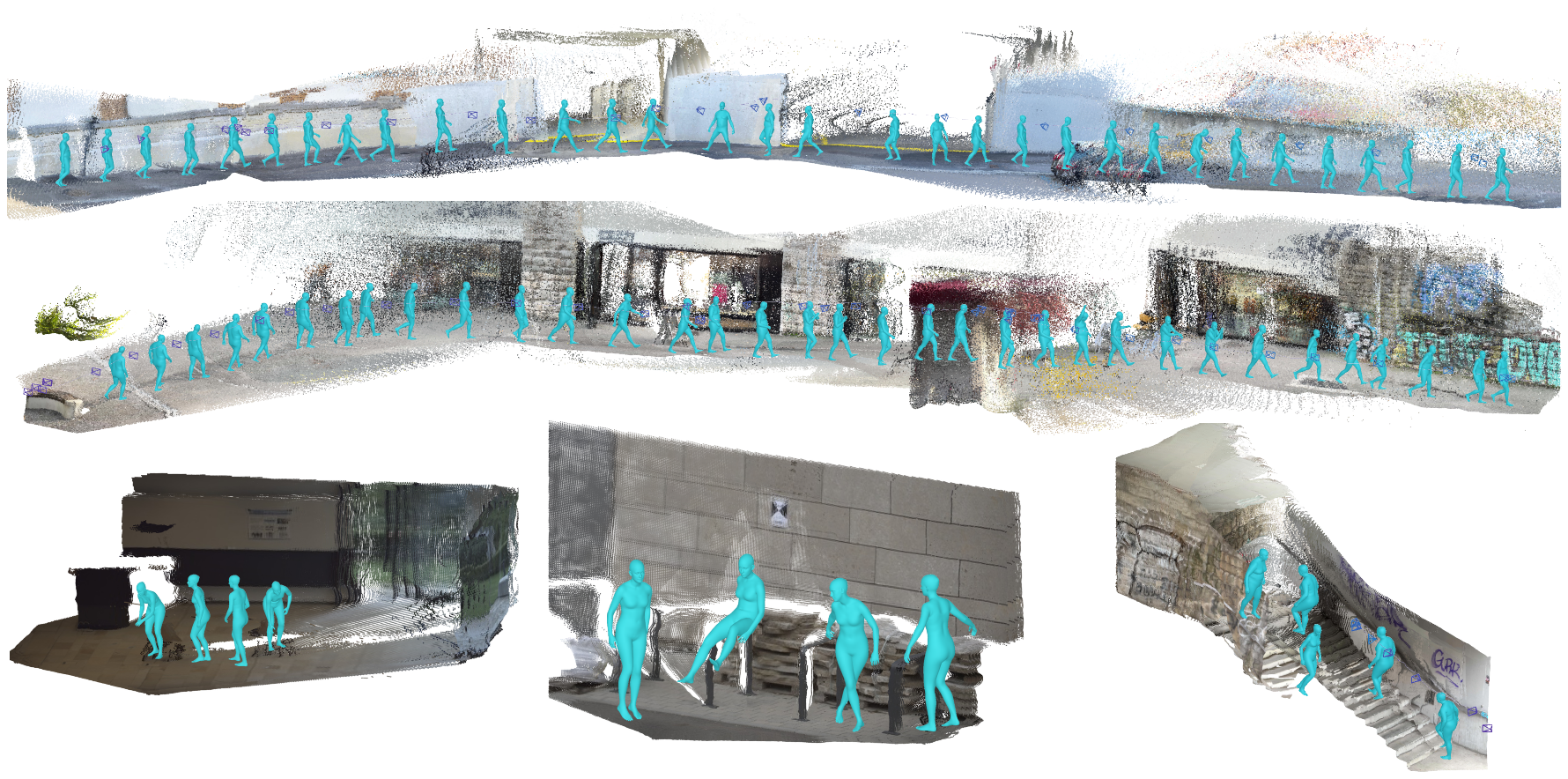}
\caption{\textbf{Qualitative 4D human--scene reconstruction.}
MoVelo jointly reconstructs human motion, scene geometry, and camera
poses online from monocular video.}
\label{fig:human-scene-reconstruction}
\end{figure*}
\endgroup

\begin{table}[tbp]
\centering
\caption{Temporal consistency on EMDB-2~\citep{kaufmann2023emdb}.
Variants A--C disable motion feature feedback; MoVelo enables it.}
\label{tab:temporal-consistency}
\papertablestyle
\begin{tabularx}{\columnwidth}{l*{3}{>{\centering\arraybackslash}X}}
\toprule
Model / Variant & \multicolumn{1}{c}{W-Jitter $\downarrow$} & \multicolumn{1}{c}{W-Accel $\downarrow$} & \multicolumn{1}{c}{W-FS $\downarrow$} \\
\midrule
Human3R & 225.1 & 46.9 & 31.6 \\
\midrule
A: Field + velocity sup. & 215.9 & 45.3 & 31.2 \\
B: A + position sup. & 216.3 & 45.4 & 31.1 \\
C: B + acceleration sup. & \underline{213.3} & \underline{44.8} & \underline{30.7} \\
\midrule
\rowcolor{tablehighlight}
\textbf{MoVelo} & \textbf{202.8} & \textbf{42.9} & \textbf{29.5} \\
\bottomrule
\end{tabularx}
\end{table}

\paragraph{Qualitative human--scene reconstruction.}
Compared with Human3R, MoVelo more closely matches the ground-truth body
positions and orientations on EMDB-2 (Fig.~\ref{fig:world-reconstruction-emdb2}).
Fig.~\ref{fig:human-scene-reconstruction} shows MoVelo's qualitative human--scene
reconstructions, including human meshes, scene point clouds, and camera poses
in a shared coordinate system.
The examples span walking, bending, raised-leg poses, and stair traversal,
illustrating articulated human motion within its scene context.
These visual results are consistent with the improvements in global joint
accuracy and temporal consistency reported in
Tables~\ref{tab:global-motion} and~\ref{tab:temporal-consistency}.
These reconstructions can also provide a coherent initialization for
subsequent human--scene optimization. Further refinement could build on
the estimated human motion, scene geometry, and camera poses to enforce
additional geometric and contact constraints.
For more visual examples, see
Section~B of Sup.Mat.

\subsection{Ablation Studies and Analysis}
\label{sec:ablations}

All ablations use EMDB-2, with other settings fixed within each comparison.
Table~\ref{tab:motion-components} compares Human3R, direct motion
regression, and our conditional velocity field. Human3R serves as the
baseline without our explicit motion modeling or feedback. The feedback
variants---decoder attention only, loss-based feedback, and motion feature
feedback---share the same conditional field and supervision.
Additional field-supervision analyses are provided in
Section~C of Sup.Mat.

\begin{table}[tbp]
\centering
\caption{Motion modeling and feedback ablations on EMDB-2.}
\label{tab:motion-components}
\papertablestyle
\normalsize
\setlength{\tabcolsep}{2pt}
\renewcommand{\arraystretch}{1.08}
\renewcommand{\tabularxcolumn}[1]{m{#1}}
\begin{tabularx}{\columnwidth}{>{\raggedright\arraybackslash}X*{5}{c}}
\toprule
Variant & \multicolumn{1}{c}{\shortstack{WA-\\MPJPE}} & \multicolumn{1}{c}{\shortstack{W-\\MPJPE}} & \multicolumn{1}{c}{\shortstack{W-\\Jitter}} & \multicolumn{1}{c}{\shortstack{W-\\Accel}} & \multicolumn{1}{c}{\shortstack{W-\\FS}} \\
\midrule
\multicolumn{6}{l}{\textit{Motion modeling}} \\
Human3R & 112.2 & 267.9 & 225.1 & 46.9 & 31.6 \\
Direct motion regression & \underline{108.4} & \underline{262.9} &
\underline{209.3} & \underline{44.1} & \underline{30.2} \\
\rowcolor{tablehighlight}
\textbf{Conditional velocity field (ours)} & \textbf{107.7} & \textbf{257.9} & \textbf{202.8} & \textbf{42.9} & \textbf{29.5} \\
\midrule
\multicolumn{6}{l}{\textit{Motion feedback}} \\
Decoder attention only & 111.3 & 266.2 & 213.3 & 44.8 & 30.7 \\
Loss-based feedback & \underline{111.0} & \underline{265.5} & \underline{209.9} & \underline{44.2} & \underline{30.3} \\
\rowcolor{tablehighlight}
\textbf{Motion feature feedback (ours)} & \textbf{107.7} & \textbf{257.9} & \textbf{202.8} & \textbf{42.9} & \textbf{29.5} \\
\bottomrule
\end{tabularx}
\end{table}

\Needspace{3\baselineskip}
\paragraph{Temporal consistency.}
\label{sec:exp-temporal}

Table~\ref{tab:temporal-consistency} examines how field supervision and
motion feature feedback contribute to temporal consistency on EMDB-2.
Variants A--C disable motion feature feedback. A uses velocity supervision,
B adds joint-position supervision, and C matches the decoder-attention-only
variant in Table~\ref{tab:motion-components}. Compared with Human3R,
the velocity-supervised field (A) reduces W-Jitter from 225.1 to 215.9 and
improves both W-Accel and W-FS, showing that learned motion cues benefit
reconstruction even without explicit feedback. Adding joint-position
supervision (B) leaves the temporal metrics nearly unchanged, suggesting
that aligning integrated positions alone provides limited additional
control over temporal fluctuations. With all three supervision terms,
variant C reduces W-Jitter to 213.3 and improves W-Accel and W-FS over B.
Finally, enabling motion feature feedback further lowers W-Jitter,
W-Accel, and W-FS to 202.8, 42.9, and 29.5, respectively. These results
suggest complementary benefits from supervising motion dynamics and
explicitly incorporating the estimated transition into human
representations before mesh regression.

\paragraph{Effect of the conditional velocity field.}
We investigate two variants of our framework for modeling motion cues:
direct motion regression and a conditional velocity field. The
direct-regression variant predicts joint velocities and accelerations
separately from the decoded temporal token. On EMDB-2, it reduces W-MPJPE
from 267.9\,mm for Human3R to 262.9\,mm
(Table~\ref{tab:motion-components}), indicating the benefit of introducing
explicit motion cues into reconstruction. The velocity-field variant
further conditions instantaneous velocity on the evolving joint state,
elapsed time, and frame interval, together with the temporal token.
Integration yields joint positions, while differentiation along the
predicted motion yields accelerations, coupling these quantities through
a shared motion function. This design further reduces W-MPJPE to
257.9\,mm and W-Jitter from 209.3 to 202.8, with lower W-Accel and W-FS.
These results support explicit motion modeling for world-grounded human
motion reconstruction and show the additional benefit of coupling
body-state transitions through a conditional velocity field.

\paragraph{Effect of motion feature feedback.}
Within our framework, the decoder-attention-only variant allows temporal tokens to
interact with human tokens through shared decoder attention. This reduces
W-MPJPE from 267.9\,mm for Human3R to 266.2\,mm on
EMDB-2 (Table~\ref{tab:motion-components}). The modest improvement may reflect
that attention exchanges motion-related information implicitly, while the
field's estimated body-state transition is not explicitly fed back into
reconstruction. Loss-based feedback adds a consistency loss between
field-integrated and reconstructed joint positions, further reducing
W-MPJPE to 265.5\,mm. Although this encourages agreement between the two
predictions during training, it does not directly incorporate the estimated
motion into the human representations used for mesh prediction. Our motion
feature feedback encodes predicted velocities and integrated displacements
to refine human representations through residual connections before mesh
regression, achieving the lowest W-MPJPE of 257.9\,mm. This refinement gives
the reconstruction head direct access to the estimated body-state
transition, allowing motion cues and current visual evidence to jointly
guide mesh prediction. This provides a more direct route for exploiting
motion estimates and may explain the larger improvement over attention-only
interaction and loss-based feedback.

\begin{table}[H]
\centering
\caption{Effect of training context length on EMDB-2.
Both runs use the same number of sampled clips and optimizer updates.}
\label{tab:training-context-length}
\papertablestyle
\normalsize
\setlength{\tabcolsep}{2pt}
\renewcommand{\arraystretch}{1.08}
\begin{tabular*}{\columnwidth}{@{\extracolsep{\fill}}c*{6}{c}}
\toprule
Training & WA- & W- & W- & W- & W- & \multirow{2}{*}{RTE} \\
clip & MPJPE & MPJPE & Jitter & Accel & FS & \\
\midrule
4 frames & \underline{107.7} & \textbf{257.9} & \underline{202.8} & \underline{42.9} & \underline{29.5} & \textbf{2.1} \\
8 frames & \textbf{106.1} & \underline{260.0} & \textbf{201.2} & \textbf{42.7} & \textbf{29.2} & \textbf{2.1} \\
\bottomrule
\end{tabular*}
\end{table}

\paragraph{Effect of training context length.}
\label{sec:exp-context-length}
\mbox{Table~\ref{tab:training-context-length}} compares four- and eight-frame training
clips with the same number of sampled clips and optimizer updates.
Longer clips improve WA-MPJPE from 107.7 to 106.1\,mm and slightly improve
the temporal metrics, but increase W-MPJPE from 257.9 to 260.0\,mm, with
unchanged RTE. These mixed results suggest that longer training context
improves temporal regularity more consistently than global localization.
We therefore use four-frame clips, reducing the number of frames processed
during training by half while maintaining comparable reconstruction
accuracy.
\section{Conclusion}
\label{sec:conclusion}

\begingroup
\brokenpenalty=10000
We present MoVelo for online, world-grounded human motion reconstruction
from monocular video. Its conditional velocity field links successive body
states in world coordinates, coupling joint positions, velocities, and
accelerations through integration, differentiation, and joint supervision.
Motion feature feedback uses predicted velocities and displacements to
refine human representations before mesh regression, combining motion cues
with current visual evidence. Experiments on EMDB-2 and RICH show improved
global motion accuracy and temporal consistency over Human3R. Ablations
further demonstrate the benefits of explicit body-state modeling and motion
feature feedback for coherent human motion reconstruction.
\par
\endgroup
\FloatBarrier
\label{main-content-end}

\Needspace{3\baselineskip}
\section*{LLM Usage Statement}
We used large language models to assist with writing and editing this paper.
These tools helped us improve the clarity and presentation of the manuscript.
We reviewed the revised text and take responsibility for the final content.

\clearpage
{\small
\renewcommand{\bibfont}{\fontsize{9}{10.55}\selectfont\interlinepenalty=10000\relax}
\setlength{\bibsep}{.85pt plus .3pt minus .2pt}
\clubpenalty=150
\widowpenalty=150
\nocite{dwivedi2024tokenhmr,li2022cliff,patel2025camerahmr,sarandi2024nlf,sun2022bev,von2018recovering,wang2025prompthmr}

\pagebreak
}

\end{document}


\flushbottom
\clubpenalty=3000
\widowpenalty=3000
\displaywidowpenalty=3000
\setcounter{page}{1}
\renewcommand{\thepage}{S\arabic{page}}
\renewcommand{\theequation}{A\arabic{equation}}
\renewcommand{\theHequation}{appendix.\arabic{equation}}
\maketitlesupplementary
\appendix
\begingroup
\setlength{\parskip}{0pt plus 1pt}
\setlength{\floatsep}{6pt plus 1pt minus 1pt}
\setlength{\textfloatsep}{8pt plus 1pt minus 1pt}
\setlength{\intextsep}{5pt plus 1pt minus 1pt}
\setlength{\dbltextfloatsep}{9pt plus 2pt minus 1pt}
\setlength{\dblfloatsep}{7pt plus 2pt minus 1pt}
\setlength{\abovecaptionskip}{4pt}
\setlength{\belowcaptionskip}{0pt}

\setcounter{dbltopnumber}{1}
\setcounter{figure}{0}
\setcounter{table}{0}
\renewcommand{\thefigure}{A\arabic{figure}}
\renewcommand{\thetable}{A\arabic{table}}
\renewcommand{\theHfigure}{appendix.\arabic{figure}}
\renewcommand{\theHtable}{appendix.\arabic{table}}

\label{sec:appendix}
\Needspace{3\baselineskip}
\section{Implementation and Training Details}
\label{sec:app-implementation}

\subsection{Network Architecture and Initialization}
\label{sec:app-network}
We compute the temporal prompt from a detached copy of the human prompt.
Both enter the shared decoder at the same image location.
Fig.~\ref{fig:app-cross-attention} shows their cross-attention maps, and
Table~\ref{tab:app-network-widths} lists the MLP widths for MoVelo-Final.
\begin{table}[H]
\centering
\caption{MLP layer widths in MoVelo-Final. Activations act between linear
layers. The query-time, frame-interval, and feedback-interval encoders have
separate weights.}
\label{tab:app-network-widths}
\papertablestyle
\begin{tabularx}{\linewidth}{@{}>{\raggedright\arraybackslash}Xcl@{}}
\toprule
Module & Layer widths & Activation \\
\midrule
Temporal prompt & $768\to768\to768$ & GELU \\
Pelvis-relative pose encoder & $66\to512\to128$ & SiLU \\
Time encoders & $1\to16\to16$ & SiLU \\
Per-joint velocity field & $966\to512\to512\to3$ & SiLU \\
Field-to-token projection & $214\to768\to768$ & GELU \\
Human feature update & $1792\to768\to1792$ & GELU \\
\bottomrule
\end{tabularx}
\end{table}

For each of the 22 SMPL-X body joints, the velocity field takes a
966-channel input (main paper, Eq.~1). It combines
the decoded temporal token (768 channels), a joint embedding (32), the
world-space pelvis position and pelvis-relative joint position (3 each),
the encoded whole-body pose (128), and separate embeddings for the query
time and frame interval (16 each). We convert the three predicted velocity
components per joint from standardized units to m/s before integration
(main paper, Eq.~3).

The feedback projection combines two standardized velocity estimates and
the integrated displacement in meters (66 channels each after flattening)
with a 16-channel frame-interval embedding. The field-derived feature updates
the temporal token only when the person matches a track in the preceding frame and
$\Delta_t>0$ (main paper, Eq.~5). For an
unmatched track, the human feature update takes the decoded temporal token
and the 1,024-channel human-prior feature. We initialize the output layers
of both feedback MLPs with zero biases and zero-mean Gaussian weights
(standard deviation $10^{-4}$). To initialize the next transition, we
transform the final SMPL-X mesh joints into the world frame using the
predicted camera pose and detach the result (main paper,
Eq.~7).

\begin{figure*}[t]
\centering
\includegraphics[width=.84\linewidth]{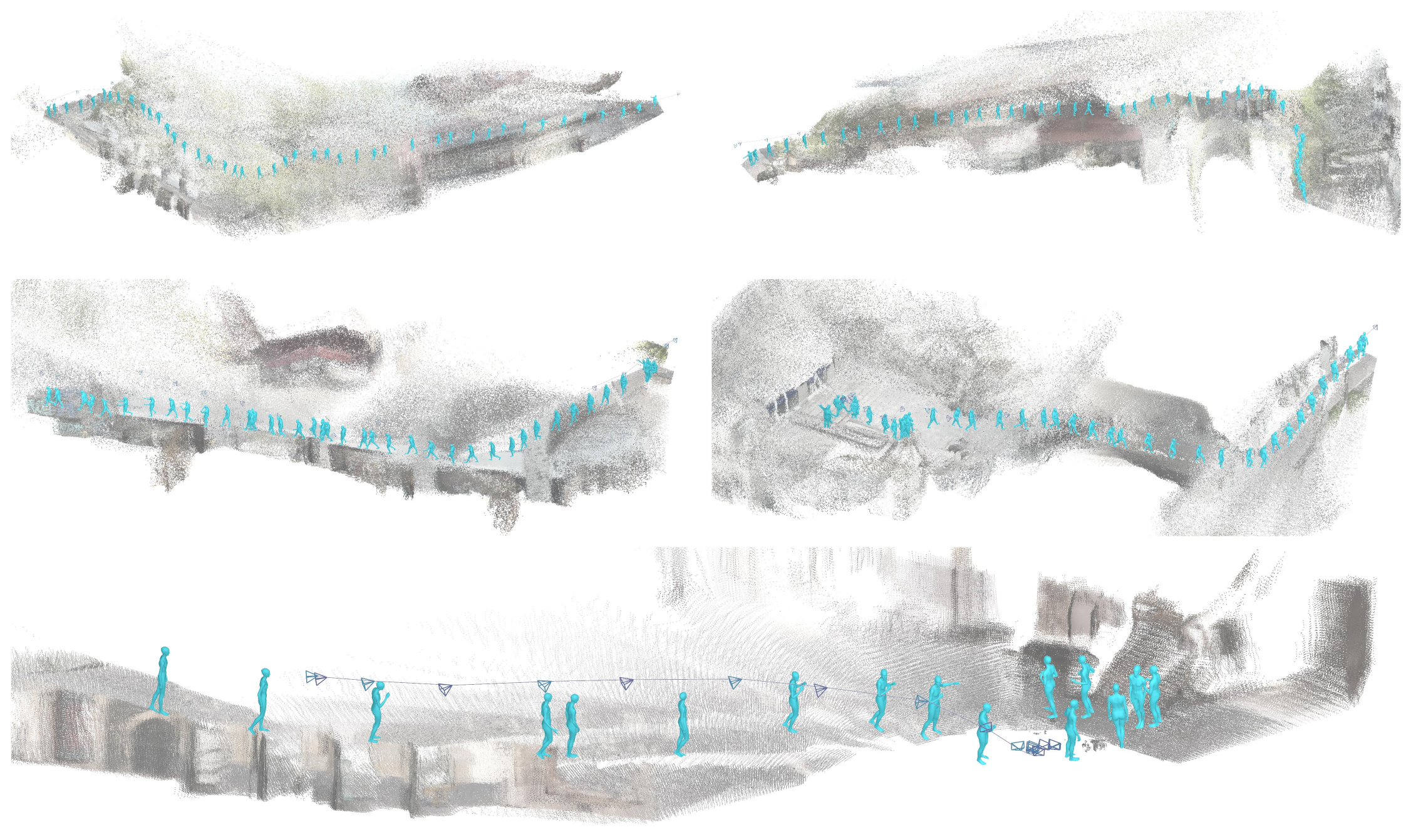}
\caption{\textbf{Long-sequence human--scene reconstruction.} We show human
meshes (cyan), the scene point cloud, and the camera
trajectory in a shared 3D coordinate system anchored to the first camera pose
(main paper, Fig.~5).}
\label{fig:app-world-scene}
\end{figure*}
\begin{figure*}[t]
\centering
\begin{minipage}[t]{.485\linewidth}
\vspace{0pt}\centering
\captionsetup{type=figure}
\includegraphics[width=.88\linewidth]{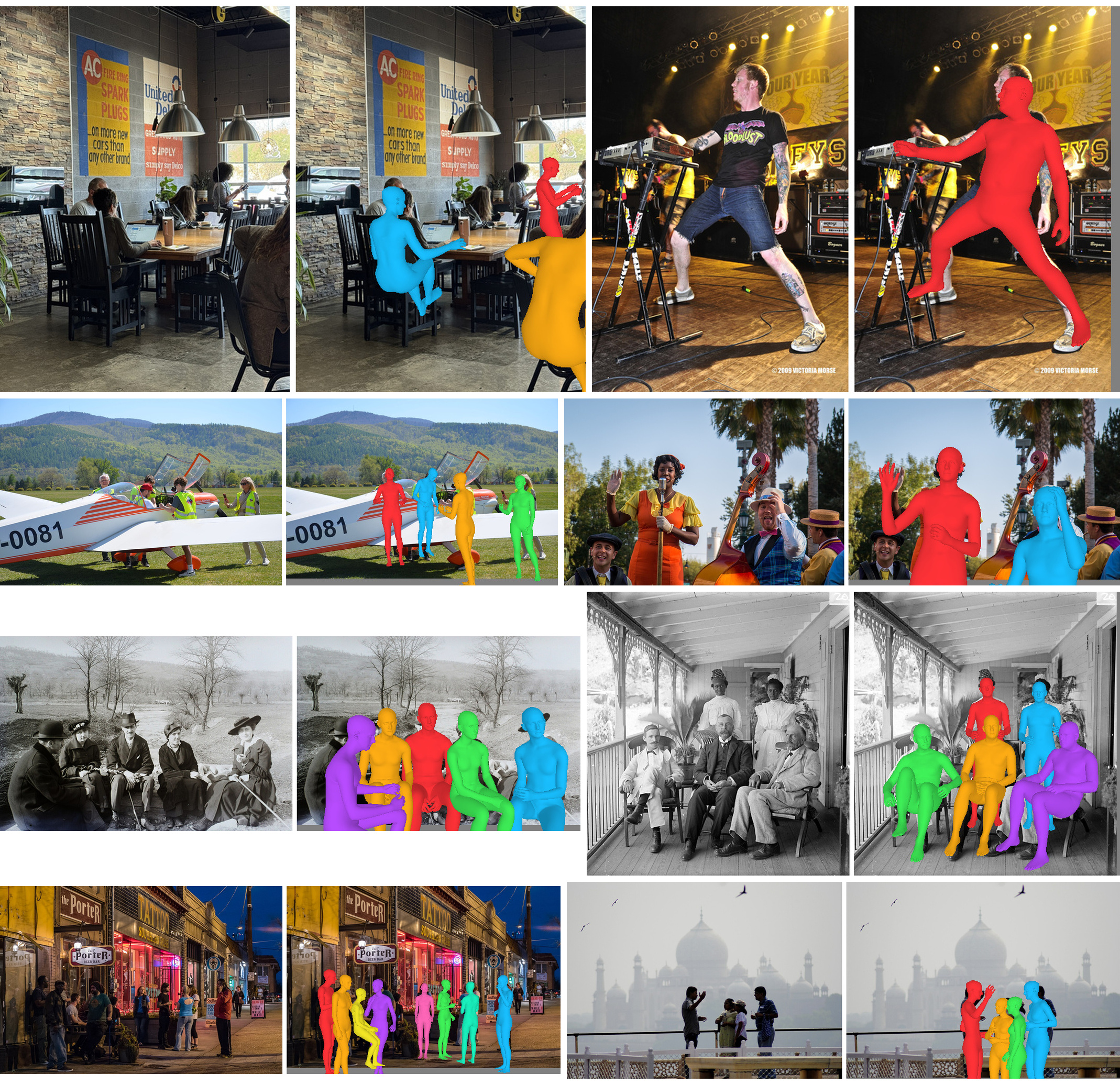}
\caption{\textbf{Occlusion and truncation.} Each input frame is paired with
MoVelo's mesh overlay for people partially hidden by others or cropped at
the image boundary.}
\label{fig:occlusion-and-truncation}
\end{minipage}\hfill
\begin{minipage}[t]{.485\linewidth}
\vspace{0pt}\centering
\captionsetup{type=figure}
\includegraphics[width=.675\linewidth]{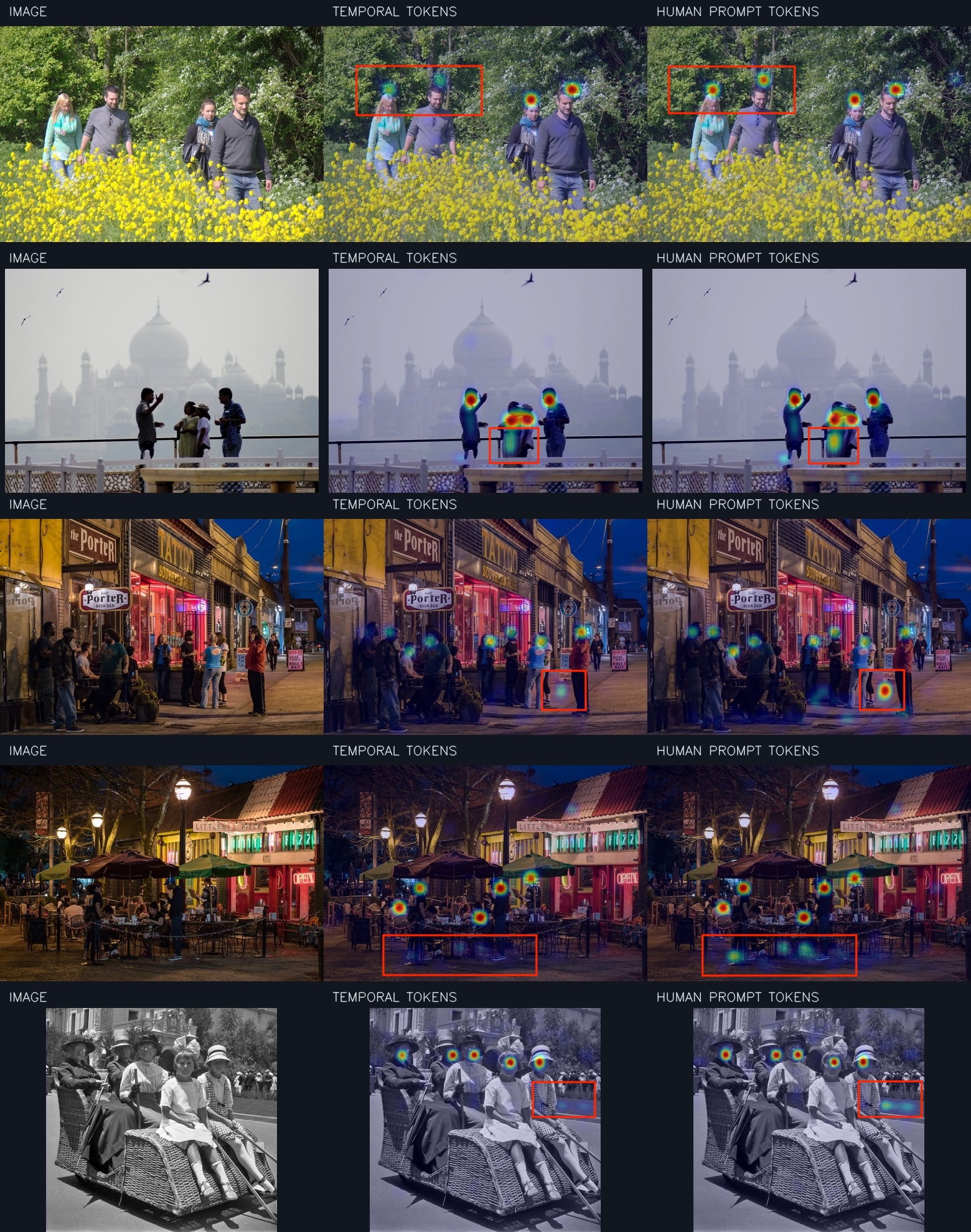}
\caption{\textbf{Decoder cross-attention.} Each row shows the input, temporal-token
attention, and human-token attention. In these examples,
temporal tokens concentrate on visible people, while human tokens also
attend to nearby scene structure.}
\label{fig:app-cross-attention}
\end{minipage}
\end{figure*}

\subsection{Training Objectives and Loss Weights}
\label{sec:app-objectives}
We minimize $\mathcal{L}_{\mathrm{recon}}+\mathcal{L}_{\mathrm{human}}
+\mathcal{L}_{\mathrm{temporal}}$ (main paper, Sec.~3.4).

\paragraph{Scene and human reconstruction.}
The scene loss sums RGB squared error, confidence-weighted pointmap error
in the current and first-camera frames (log-confidence penalty $0.2$),
and Euclidean errors on normalized camera translation and rotation
quaternions. The human loss is
\begin{equation}
\begin{aligned}
\mathcal{L}_{\mathrm{human}} &=
0.1\mathcal{L}_{\mathrm{det}} + \mathcal{L}_{\mathrm{mask}} \\
&\quad + \mathcal{L}_{\mathrm{pose}} + 0.1\mathcal{L}_{\mathrm{shape}} \\
&\quad + \mathcal{L}_{\mathrm{trans}} + \mathcal{L}_{\mathrm{3D}} +
0.01\mathcal{L}_{\mathrm{2D}} .
\end{aligned}
\label{eq:human-objective}
\end{equation}
Detection uses focal loss and the mask uses binary cross-entropy. The
remaining terms use L1 errors on pose rotation matrices, shape coefficients,
head-centered translation, joints and vertices in pelvis-relative and camera
coordinates, and their image reprojections, respectively.

\paragraph{Temporal supervision.}
We express training targets in the coordinate frame of the clip's first
annotated camera. We apply the full rigid transform to positions and its
rotation to velocities and accelerations. For a person present
at both sampled frames, the interval-average velocity target is
$\hat{\mathbf{v}}_t=(\hat{\mathbf{X}}_t-\hat{\mathbf{X}}_{t-1})/\Delta_t$.
The acceleration target $\hat{\mathbf{a}}_t$ uses centered differences at
neighboring timestamps in the original sequence, so a later annotation can
contribute during training. At inference, the reconstruction at $t$ uses
only observations through $t$.

With $\ell(\mathbf{u},\hat{\mathbf{u}})=
\|\mathbf{u}-\hat{\mathbf{u}}\|_{\mathrm{F}}^2/(3J)$ and $J=22$,
we compute the temporal loss as
\begin{equation}
\begin{aligned}
\mathcal{L}_{\mathrm{temporal}} &=
\ell(\mathbf{X}^{\mathrm{field}}_t,\hat{\mathbf{X}}_t) \\
&\quad + \ell(\widetilde{\mathbf{v}}_t,\widetilde{\hat{\mathbf{v}}}_t) \\
&\quad + \ell(\widetilde{\mathbf{a}}^{\mathrm{field}}_t,\widetilde{\hat{\mathbf{a}}}_t).
\end{aligned}
\label{eq:temporal-objective}
\end{equation}
Joint positions enter the loss in meters. Tildes denote velocities (m/s)
and accelerations (m/s$^2$) standardized using a fixed training-set mean
and standard deviation for each joint coordinate. Within each frame, we
average each term over matched people with valid image and human
annotations and a valid target for that term. We then average over frames,
assigning zero to a term in frames with no valid targets.

\paragraph{Endpoint acceleration.}
Heun's step uses velocities at the starting state and at its Euler
prediction. For acceleration, we evaluate the field again at the corrected
endpoint $\mathbf{X}^{\mathrm{field}}_t$. With the temporal token
$\mathbf{z}_t$ and interval $\Delta_t$ fixed, Eq.~4
of the main paper gives
\begin{equation}
\mathbf{a}^{\mathrm{field}}_t =
\left[\partial_s f_\phi + D_{\mathbf{X}} f_\phi[f_\phi]\right]_{
\mathbf{X}=\mathbf{X}^{\mathrm{field}}_t,\;s=\Delta_t}.
\label{eq:app-total-acceleration}
\end{equation}
We compute this Jacobian--vector product in $(\mathbf{X},s)$ along
$(f_\phi,1)$, where $f_\phi$ is the physical joint velocity at that endpoint.

\subsection{Training Settings}
\label{sec:app-training}
We initialize the scene and human-prior modules from CUT3R~\citep{wang2025cut3r}
and Multi-HMR~\citep{baradel2024multihmr}. The scene encoder, shared decoder,
human-prior encoder, detector, and scene and camera heads stay frozen. We train
the human prompt and reconstruction head together with the temporal prompt,
conditional velocity field, and motion feature feedback networks.

Each epoch samples 4,000 four-frame clips from the filtered BEDLAM training
split~\citep{black2023bedlam}. Successive sampled frames are at most four
indices apart. Batches of eight clips give 500 updates per epoch and 20,000
over 40 epochs. The eight-frame run doubles the number of training frames
at the same clip and update counts (main paper,
Table~5). AdamW uses
$\beta=(0.9,0.95)$, weight decay $0.05$, and gradient-norm clipping at $1$.
The learning rate warms up to $5\times10^{-5}$ over five epochs, then decays
on a cosine schedule toward $10^{-6}$. The training seed is 0 and the
per-epoch sampling seed is $777+\mathrm{epoch}$. Training 73.23M parameters
(6.24\% of the model) takes approximately one day on a single RTX 4090.
The measured inference cost of the final model is $5.734\times10^{12}$ FLOPs.

\Needspace{3\baselineskip}
\subsection{Evaluation Protocols and Metrics}
\label{sec:app-evaluation}
We process all 43,129 frames of the 25 EMDB-2 sequences~\citep{kaufmann2023emdb}
and all frames of the 191 static-camera RICH sequences~\citep{huang2022capturing}
in order, with the long side of the scene input resized to 512 pixels.
Recurrent state resets every 50 frames on EMDB-2 and carries through each
full RICH sequence.

\paragraph{Global motion.}
W-MPJPE and WA-MPJPE measure mean joint-position error in millimeters over
segments of up to 100 frames. For W-MPJPE$_{100}$, we fit a similarity
transform to the first two frames and apply it to the whole segment.
WA-MPJPE fits the transform to all frames in the segment. RTE measures
mean root-position error after rigid trajectory alignment as a percentage
of the ground-truth path length.

\paragraph{Temporal consistency.}
W-Jitter is the mean norm of the third finite difference of predicted
world-space joints, scaled by $\mathrm{FPS}^3/10$. W-Accel compares second
finite differences of predicted and annotated joints over consecutive
frames of the same person. It retains scale and root motion in the first
evaluated camera's orientation and weights results by valid window counts.
W-FS averages the predicted interframe displacement (mm) of four foot
vertices, counting a vertex whenever its ground-truth displacement is below 10\,mm.

\section{Additional Visualizations}
\label{sec:app-qualitative}
In the three walking sequences shown in Fig.~\ref{fig:app-emdb2-trajectories},
MoVelo's trajectories align more closely with the ground truth than Human3R's.

\section{Additional Experimental Results}
\label{sec:app-quantitative}

\subsection{Velocity Field Supervision}
\label{sec:app-supervision-ablation}
With position, velocity, and acceleration supervision, motion feature
feedback lowers W-MPJPE on EMDB-2 from 266.2\,mm for variant C to
257.9\,mm for MoVelo (Table~\ref{tab:field-supervision}). With feedback
enabled, supervising acceleration in addition to joint positions and
velocities reduces W-MPJPE by 1.8\,mm.
\begin{table}[htbp]
\centering
\caption{Velocity field supervision on EMDB-2. In variants \mbox{A--C}, motion cues
reach the human representation indirectly through decoder self-attention.
Bold marks the lowest error within each feedback setting.}
\label{tab:field-supervision}
\papertablestyle
\footnotesize
\renewcommand{\arraystretch}{1.28}
\setlength{\tabcolsep}{2pt}
\begin{tabularx}{\linewidth}{@{}>{\raggedright\arraybackslash}Xrrrrr@{}}
\toprule
Variant & \shortstack[r]{WA-\\MPJPE} & \shortstack[r]{W-\\MPJPE} & W-Jitter & W-Accel & W-FS \\
\midrule
Human3R & 112.2 & 267.9 & 225.1 & 46.9 & 31.6 \\
\midrule
\multicolumn{6}{l}{\textit{Motion feature feedback disabled}} \\
A: Field + velocity sup. & 111.9 & 266.9 & 215.9 & 45.3 & 31.2 \\
B: A + position sup. & 112.0 & 267.7 & 216.3 & 45.4 & 31.1 \\
C: B + acceleration sup. & \textbf{111.3} & \textbf{266.2} & \textbf{213.3} & \textbf{44.8} & \textbf{30.7} \\
\midrule
\multicolumn{6}{l}{\textit{Motion feature feedback enabled}} \\
Velocity only & 108.0 & 259.8 & 205.3 & 43.4 & 29.6 \\
Velocity + position & \textbf{107.4} & 259.7 & 203.2 & 43.0 & \textbf{29.4} \\
Velocity + acceleration & 109.0 & 264.2 & 205.8 & 43.5 & 29.6 \\
\rowcolor{tablehighlight}
\textbf{MoVelo} (all three) & 107.7 & \textbf{257.9} & \textbf{202.8} & \textbf{42.9} & 29.5 \\
\bottomrule
\end{tabularx}
\end{table}

\subsection{Loss on Integrated Joint Positions}
\label{sec:app-endpoint-objective}
Switching the loss on integrated joint positions from Smooth-L1 to MSE
lowers W-MPJPE by 1.4\,mm on EMDB-2 and 3.1\,mm on RICH
(Table~\ref{tab:joint-position-loss}). On EMDB-2, the two losses give the
same W-Accel (42.9), and Smooth-L1 gives slightly lower W-Jitter and W-FS.
On RICH, MSE yields lower errors on all five metrics.
\begin{table}[H]
\centering
\caption{Joint-position loss with motion feature feedback. Both variants use
MSE on standardized velocities and accelerations.}
\label{tab:joint-position-loss}
\papertablestyle
\footnotesize
\renewcommand{\arraystretch}{1.28}
\setlength{\tabcolsep}{2pt}
\begin{tabular*}{\linewidth}{@{\extracolsep{\fill}}llrrrrr@{}}
\toprule
Dataset & Position loss & \shortstack[r]{WA-\\MPJPE} & \shortstack[r]{W-\\MPJPE} & W-Jitter & W-Accel & W-FS \\
\midrule
EMDB-2 & Smooth-L1 & 107.8 & 259.3 & \textbf{202.6} & \textbf{42.9} & \textbf{29.3} \\
       & MSE       & \textbf{107.7} & \textbf{257.9} & 202.8 & \textbf{42.9} & 29.5 \\
\midrule
RICH   & Smooth-L1 & 105.4 & 173.8 & 149.1 & 28.1 & 23.4 \\
       & MSE       & \textbf{105.1} & \textbf{170.7} & \textbf{146.2} & \textbf{27.6} & \textbf{23.2} \\
\bottomrule
\end{tabular*}
\end{table}

\twocolumn[{
\begin{minipage}{\textwidth}
\centering
\captionsetup{type=figure}
\includegraphics[width=\linewidth]{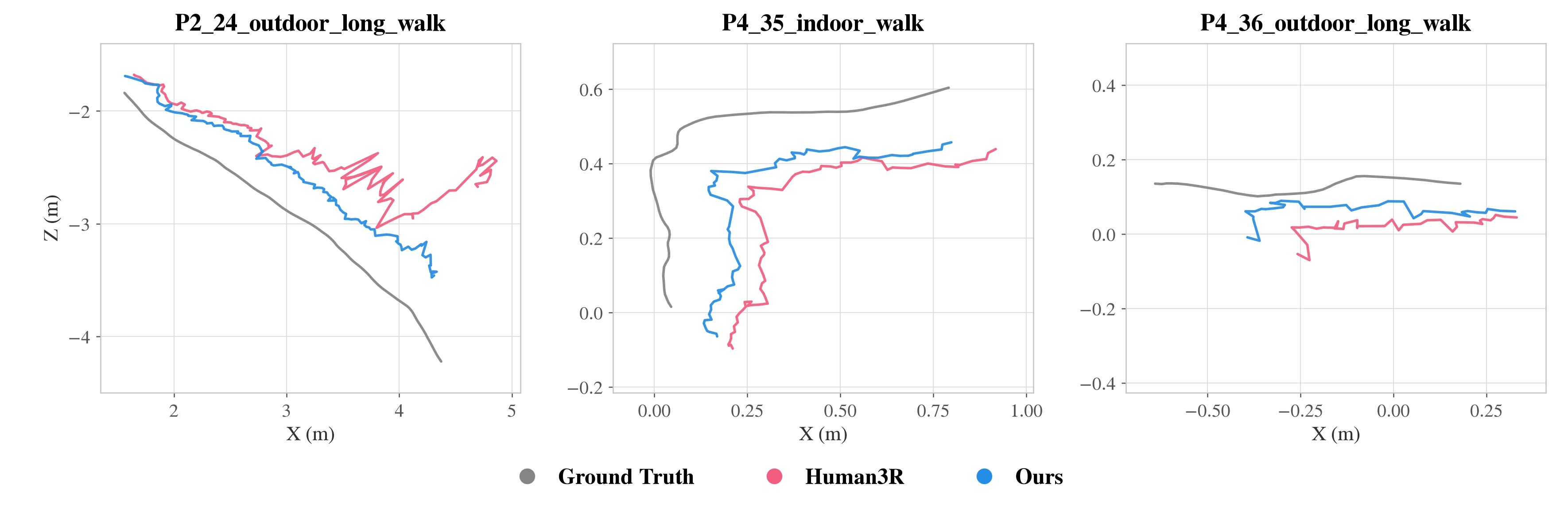}
\caption{\textbf{Human trajectories on EMDB-2.} We plot two outdoor walks
(left and right) and an indoor walk (center) in the world $X$--$Z$ plane (meters).
Gray denotes ground truth, pink Human3R, and blue MoVelo.}
\label{fig:app-emdb2-trajectories}
\end{minipage}
}]
\raggedbottom

\subsection{Local Mesh Reconstruction}
\label{sec:app-local-reconstruction}
MPJPE and PVE measure pelvis-relative joint and vertex errors. PA-MPJPE
uses Procrustes alignment. Among one-stage methods (Table~\ref{tab:local-reconstruction}),
MoVelo leads on all three 3DPW metrics, improving on Human3R by 0.7\,mm in
MPJPE and 0.5\,mm in PVE. On EMDB-1, MoVelo has the lowest PVE (85.8\,mm) in this
group.
\begin{table}[H]
\centering
\caption{Local mesh errors (mm) on 3DPW and EMDB-1. Bold and underline
mark the best and second-best results in each group. Dashes denote
unreported results.}
\label{tab:local-reconstruction}
\papertablestyle
\footnotesize
\setlength{\tabcolsep}{1.5pt}
\renewcommand{\arraystretch}{1.05}
\begin{tabularx}{\columnwidth}{@{}l*{6}{>{\raggedleft\arraybackslash}X}@{}}
\toprule
& \multicolumn{3}{c}{3DPW~\citep{von2018recovering}} & \multicolumn{3}{c}{EMDB-1~\citep{kaufmann2023emdb}} \\
\cmidrule(lr){2-4}\cmidrule(lr){5-7}
Method & \shortstack[r]{PA-\\MPJPE} & MPJPE & PVE & \shortstack[r]{PA-\\MPJPE} & MPJPE & PVE \\
\midrule
\multicolumn{7}{l}{\textit{Multi-stage}} \\
CLIFF~\citep{li2022cliff} & 43.0 & 69.0 & 81.2 & 68.3 & 103.3 & 123.7 \\
HMR 2.0a~\citep{goel2023humans4d} & 44.4 & 69.8 & 82.2 & 61.5 & 97.8 & 120.0 \\
TokenHMR~\citep{dwivedi2024tokenhmr} & 44.3 & 71.0 & 84.6 & 55.6 & 91.7 & 109.4 \\
CameraHMR~\citep{patel2025camerahmr} & 38.5 & 62.1 & 72.9 & 43.7 & 73.0 & 85.4 \\
NLF~\citep{sarandi2024nlf} & \underline{37.3} & \underline{60.3} & \underline{71.4} & \underline{41.2} & \underline{69.6} & \underline{82.4} \\
OnlineHMR~\citep{zhao2026onlinehmr} & 43.7 & 69.9 & 83.7 & 46.0 & 74.0 & 86.1 \\
PromptHMR~\citep{wang2025prompthmr} & \textbf{35.5} & \textbf{56.9} & \textbf{67.3} & \textbf{40.1} & \textbf{68.1} & \textbf{79.2} \\
\midrule
\multicolumn{7}{l}{\textit{One-stage}} \\
BEV~\citep{sun2022bev} & 46.9 & 78.5 & 92.3 & 70.9 & 112.2 & 133.4 \\
Multi-HMR~\citep{baradel2024multihmr} & 45.9 & 73.1 & 87.1 & \underline{50.1} & 81.6 & 95.7 \\
UniSH~\citep{li2026unish} & -- & 80.1 & 94.2 & -- & -- & -- \\
UniCon3R~\citep{sur2026unicon3r} & -- & 72.5 & 86.8 & -- & -- & -- \\
Human3R~\citep{chen2026human3r} & \underline{44.1} & \underline{71.2} & \underline{84.9} & \textbf{48.5} & \textbf{73.9} & \underline{86.0} \\
\rowcolor{tablehighlight}
\textbf{MoVelo} & \textbf{43.9} & \textbf{70.5} & \textbf{84.4} & 51.4 & \underline{76.4} & \textbf{85.8} \\
\bottomrule
\end{tabularx}
\end{table}

\subsection{Temporal Consistency on RICH}
\label{sec:app-rich-temporal}
On RICH, MoVelo reduces world-space jitter by 14.5\%, acceleration error by
14.0\%, and foot sliding by 10.8\% relative to Human3R
(Table~\ref{tab:app-rich-temporal}). On the same sequences, MoVelo improves
W-MPJPE from 184.9 to 170.7\,mm (main paper, Table~1).
\begin{table}[H]
\centering
\caption{Temporal consistency on the 191 static-camera RICH
sequences~\citep{huang2022capturing}.}
\label{tab:app-rich-temporal}
\papertablestyle
\begin{tabularx}{\linewidth}{l*{3}{>{\raggedleft\arraybackslash}X}}
\toprule
Method & W-Jitter $\downarrow$ & W-Accel $\downarrow$ & W-FS $\downarrow$ \\
\midrule
Human3R~\citep{chen2026human3r} & 170.9 & 32.1 & 26.0 \\
\rowcolor{tablehighlight}
\textbf{MoVelo} & \textbf{146.2} & \textbf{27.6} & \textbf{23.2} \\
\bottomrule
\end{tabularx}
\end{table}

\Needspace{3\baselineskip}
\section{Limitations}
\label{sec:app-limitations}
Mesh and camera-pose errors can shift the starting joints for the next
transition, and a missing identity match interrupts field feedback.
Robustness to prolonged occlusions and rapid camera motion remains open.
\endgroup
\FloatBarrier
\label{supp-content-end}